\documentclass[10pt]{article} % For LaTeX2e
\usepackage[accepted]{tmlr}
\usepackage{amsmath,amsfonts,bm}

\def\eqref#1{equation~\ref{#1}}
\def\1{\bm{1}}

\DeclareMathAlphabet{\mathsfit}{\encodingdefault}{\sfdefault}{m}{sl}
\SetMathAlphabet{\mathsfit}{bold}{\encodingdefault}{\sfdefault}{bx}{n}

\usepackage{hyperref}
\usepackage{url}
\usepackage{graphicx}
\usepackage{algorithm}% http://ctan.org/pkg/algorithms
\usepackage{algpseudocode}% http://ctan.org/pkg/algorithmicx

\usepackage{wrapfig}
\usepackage{multicol}
\usepackage{multirow}
\usepackage{sidecap}

\title{Towards Understanding Dual BN In Hybrid Adversarial Training}

\author{\name Chenshuang Zhang  \email zcs15@kaist.ac.kr  \\
   \addr KAIST  \\
 \AND
 \name Chaoning Zhang \thanks{Corresponding author}  \email chaoningzhang1990@gmail.com \\
   \addr  Kyung Hee University
 \AND   
 \name Kang Zhang   \email zhangkang@kaist.ac.kr\\
\addr KAIST 
 \AND 
 \name Axi Niu  \email nax@mail.nwpu.edu.cn \\
 \addr Northwestern Polytechnical University
 \AND  
  \name Junmo Kim  \email junmo.kim@kaist.ac.kr \\
  \addr KAIST
  \AND 
    \name In So Kweon  \email iskweon77@kaist.ac.kr \\
\addr KAIST
}

\def\month{03}  % Insert correct month for camera-ready version
\def\year{2024} % Insert correct year for camera-ready version
\def\openreview{\url{https://openreview.net/forum?id=bQKHMSE4SH}} % Insert correct link to OpenReview for camera-ready version

\begin{document}

\maketitle

\begin{abstract}
There is a growing concern about applying batch normalization (BN) in adversarial training (AT), especially when the model is trained on both \textit{adversarial} samples and \textit{clean} samples (termed Hybrid-AT). With the assumption that \textit{adversarial} and \textit{clean} samples are from two different domains, a common practice in prior works is to adopt Dual BN, where BN$_{adv}$ and BN$_{clean}$ are used for adversarial and clean branches, respectively. A popular belief for motivating Dual BN is that estimating normalization statistics of this mixture distribution is challenging and thus disentangling it for normalization achieves stronger robustness. In contrast to this belief, we reveal that disentangling statistics plays a less role than disentangling affine parameters in model training. This finding aligns with prior work  ~\citep{rebuffi2023revisiting}, and we build upon their research for further investigations. We demonstrate that the domain gap between adversarial and clean samples is not very large, which is counter-intuitive considering the significant influence of adversarial perturbation on the model accuracy. We further propose a two-task hypothesis which serves as the empirical foundation and a unified framework for Hybrid-AT improvement. We also investigate Dual BN in test-time and reveal that affine parameters characterize the robustness during inference. Overall, our work sheds new light on understanding the mechanism of Dual BN in Hybrid-AT and its underlying justification. 
\end{abstract}

\section{Introduction}
\label{sec:intro}

Adversarial training (AT)~\citep{ganin2016domain,madry2017towards,shafahi2019adversarial,andriushchenko2020understanding,bai2021recent} that optimizes the model on adversarial examples is a time-tested and effective technique for improving robustness against adversarial attack~\citep{qiu2019review,xu2020adversarial,dong2018boosting,zhang2021survey}. Beyond classical AT (also termed Madry-AT)~\citep{madry2017towards}, a common AT setup is to train the model on both \textit{adversarial} samples and \textit{clean} samples  (termed Hybrid-AT)~\citep{goodfellow2014explaining,kannan2018adversarial,xie2019intriguing,xie2020adversarial}. Batch normalization (BN)~\citep{ioffe2015batch,santurkar2018does,bjorck2018understanding,li2016revisiting} has become a de facto standard component in modern deep neural networks (DNNs)~\citep{he2016deep,huang2017densely,zhang2019revisiting,zhang2020resnet}, however, there is a notable concern regarding how to use BN in the Hybrid-AT setup. This concern mainly stems from ~\citet{xie2019intriguing,xie2020adversarial}, which claims the adversarial and clean samples are from two different domains, and thus a separate BN should be used for each domain. This technique applying different BN for different domains has been adopted in multiple works with different names, e.g., Dual BN~\citep{jiang2020robust,wang2020once,wang2021augmax} and mixture BN~\citep{xie2019intriguing}. With different names, however, they refer to the same practice of adopting BN$_{adv}$ and BN$_{clean}$ for adversarial and clean samples, respectively. To avoid confusion, we use Dual BN for the remainder of this work.

Despite the increasing popularity of Dual BN, the mechanism of how Dual BN helps Hybrid-AT remains not fully clear. Toward a better understanding of this mechanism, we revisit a long-held belief in~\citet{xie2019intriguing,xie2020adversarial}. Specifically, it justifies the necessity of Dual BN in Hybrid-AT with the following claim (quoted from the abstract of~\citet{xie2019intriguing}):

\textit{``Estimating normalization statistics of the mixture distribution is challenging"} and \textit{``disentangling the
mixture distribution for normalization, i.e., applying separate BNs to clean and adversarial images for statistics estimation, achieves much stronger robustness."} 

The above claim~\citep{xie2019intriguing} emphasizes the necessity of disentangling the normalization statistics (NS) in Hybrid-AT. The underlying motivation for the above claim is that BN statistics calculated on the clean domain are incompatible with training the model on the adversarial domain, and vice versa. Therefore, Hybrid-AT with a single BN suffers from such incompatibility with BN statistics calculated from the mixed distribution, while Dual BN can avoid the incompatibility through training the clean and adversarial samples with two BN branches separately. As a preliminary investigation, our work experiments with a new variant of AT with Cross-BN, namely training the adversarial samples with BN$_{clean}$ and vice versa. Interestingly, we find that using BN from another domain only has limited influence on the performance. This observation inspires us to have a closer look at how Dual BN works in Hybrid-AT. Through untwining normalization statistics (NS) and affine parameters (AP) in Dual BN to include one effect while excluding the other, we demonstrate that two AP sets can achieve comparable performance to the original Dual BN, which is consistent with the finding in ~\citet{rebuffi2023revisiting}. We also reveal that disentangled NS can achieve similar performance to Dual BN under certain conditions like small perturbations ($\epsilon=8/255$). These findings refute the prior claim emphasizing the role of disentangled NS in Dual BN~\citep{xie2019intriguing,xie2020adversarial}, and also inspires us to investigate whether the motivation for Dual BN holds, i.e., the two-domain hypothesis in ~\citet{xie2019intriguing,xie2020adversarial}.

As the motivation for adopting Dual BN, the two-domain hypothesis assumes that \textit{``clean images and adversarial images are drawn from two different domains"}  (quoted from ~\citet{xie2019intriguing}). This hypothesis is verified in~\citet{xie2019intriguing} mainly by the visualization of NS, which highlights a large adversarial-clean domain gap. However, we point out that their visualization has a hidden flaw, which makes their claim regarding the domain gap between adversarial and clean samples deserve a closer look.
Specifically, the visualization in~\citet{xie2019intriguing} ignores the influence of different AP when calculating NS. After fixing this hidden flaw, we demonstrate that the adversarial-clean domain gap is not as large as claimed in prior work. Interestingly, under the same perturbation/noise magnitude, we show that there is no significant difference between adversarial-clean domain gap and noisy-clean counterpart. 

Inspired by the above findings, we propose a two-task hypothesis to replace the two-domain hypothesis in ~\citet{xie2019intriguing,xie2020adversarial} for justification on how Dual BN works in Hybrid-AT. Specifically, we claim that there are two tasks in Hybrid-AT: one task for clean accuracy and the other for robustness. Our two-task hypothesis offers empirical foundations and a unified framework for Hybrid-AT improvements, which generalizes Hybrid-AT with Dual BN to various model designs, including the adapter method in ~\citet{rebuffi2023revisiting} and Trades-AT~\citep{zhang2019theoretically}. In addition to exploring BN for training Hybrid-AT models, our study delves into Dual BN at test time, uncovering that affine parameters characterize robustness during inference.

We summarize our main contributions as follows.

\begin{itemize}

    \item Our work thoroughly investigates how disentangled normalization statistics (NS) and affine parameters (AP) in Dual BN impact the training Hybrid-AT models, leading to a comprehensive and solid refutation of prior claims about the significance of NS.

    \item Our work investigates the adversarial-clean domain gap.  We point out a hidden flaw of NS visualization in prior work, and demonstrate the adversarial-clean domain gap is not as large as expected both visually and quantitatively.

    \item Our work proposes a two-task hypothesis as an empirical foundation and unified framework for enhancing Hybrid-AT, connecting diverse methods like Dual BN, Dual Linear, Adapters and Trades-AT. This hypothesis may bring new inspirations to Hybrid-AT improvements from a new perspective.

    \item Our study examines Dual BN at test time, exploring various NS and AP types with a pretrained model and revealing that AP determines robustness during inference.

\end{itemize}

\section{Problem overview}
\label{sec:problem_overview}

\subsection{Adversarial training}
\label{sec:problem_overview_at}

\textbf{Adversarial training.} Adversarial training (AT)~\citep{ganin2016domain,madry2017towards,shafahi2019adversarial,andriushchenko2020understanding,bai2021recent} has been the most powerful defense method against adversarial attacks, among which Madry-AT~\citep{madry2017towards} is a typical method detailed as follows. Let's assume $\mathcal{D}$ is a data distribution with $(x, y)$ pairs and $f(\cdot ,  \theta)$ is a model parametrized by $\theta$. $l$ indicates cross-entropy loss in classification. Instead of directly feeding clean samples from $\mathcal{D}$ to minimize the risk of $ \mathbb{E}_{(x,y) \sim \mathcal{D}} [l(f(x,\theta), y)]$, ~\citet{madry2017towards} formulates a saddle problem for finding model parameter $\theta$ by optimizing the following adversarial risk:
\begin{equation}
\arg\min_{\mathbf{\theta}} \mathbb{E}_{(x,y) \sim \mathcal{D}}\left[ \max_{\delta \in \mathbb{S}} l(f(x + \delta; \theta), y) \right]
\label{eq:madry}
\end{equation}
where $\mathbb{S}$ denotes the allowed perturbation budget which is a typically $l_p$ norm-bounded $\epsilon$. We term the above adversarial training framework as Classical-AT. It adopts a two-step training procedure (inner maximization + outer minimization), and trains the robust model with only adversarial samples. Following the same procedure, ~\citet{xie2019intriguing,xie2020adversarial} propose to train the robust model with both clean and adversarial samples, termed as \textbf{Hybrid-AT}. The loss of Hybrid-AT is defined as follows: 
\begin{equation}
     \mathcal{L}_{Hybrid} = \alpha l(f(x; \theta), y) + (1 - \alpha)  l(f(x + \delta; \theta), y) 
     \label{eq:hybrid}
\end{equation}
where $x$ and $x + \delta$ indicate clean and adversarial samples, respectively.
$\alpha$ is a hyper-parameter for balancing the clean and adversarial branches, is set to 0.5 in this work following~\citet{goodfellow2014explaining,xie2019intriguing}.

\subsection{Batch normalization in AT}
\label{sec:problem_overview_bn}

\textbf{Batch normalization (BN).}
We briefly summarize how BN works in modern networks. For a certain layer in the DNN, we denote the feature layers of a mini-batch in the DNN as $\mathcal{B} = \{x^1, ..., x^m\}$. The feature layers are normalized by mean $\mu$ and standard deviation $\sigma$ as:
\begin{equation}
     \hat{x}^i = \frac{x^i - \mu}{\sigma} \cdot \gamma + \beta
     \label{eq:BN_principle}
\end{equation}
where $\gamma$ and $\beta$ indicate the weight and bias in BN, respectively. To be clear, we refer $\mu$ and $\sigma$ as normalization statistics (NS),  $\gamma$ and $\beta$ as affine parameters (AP). During training, NS is calculated on the current mini-batch statistics for the update of model weights. Meanwhile, a running average of NS is recorded in the whole training process, which is applied for inference after training ends.

\textbf{Dual BN in AT.} There is an increasing interest in investigating BN in the context of adversarial robustness~\citep{awais2020towards,cheng2020adversarial,nandy2021adversarially,sitawarinimproving,gong2022sandwich}. This work focuses on Hybrid-AT with Dual BN~\citep{xie2019intriguing,xie2020adversarial} which applies BN$_{clean}$ and BN$_{adv}$ to clean branch and adversarial branch, respectively.

\subsection{Experimental setups}
~\citet{pang2020bag} demonstrates that AT's basic training settings significantly impact model performance and recommended specific parameters for a fair comparison of AT methods. If not specified, we adhere to the suggested settings in ~\citet{pang2020bag}.

\textbf{Experimental setups.} In this work, we perform experiments on CIFAR10~\citep{krizhevsky2009learning,andriushchenko2020understanding,zhang2022decoupled} with ResNet18~\citep{andriushchenko2020understanding,targ2016resnet,wu2019wider,li2016demystifying,zhang2022decoupled}. Specifically, we train the model for 110 epochs. The learning rate is set to 0.1 and decays by a factor of 0.1 at the epoch 100 and 105. We adopt an SGD optimizer with weight decay $5\times10^{-4}$. For generating adversarial examples during training, we use $\ell_{\infty}$ PGD attack with 10 iterations and step size $\alpha=2/255$. For the perturbation constraint, $\epsilon$ is set to $\ell_{\infty}$ $8/255$~\citep{pang2020bag} or $16/255$~\citep{xie2019intriguing}. Following~\citet{pang2020bag}, we evaluate the model robustness under PGD-10 attack (PGD attack with 10 steps) and AutoAttack (AA)~\citep{croce2020reliable}. 

\begin{figure}[tbp]
  \centering
    \includegraphics[width=0.7\linewidth]{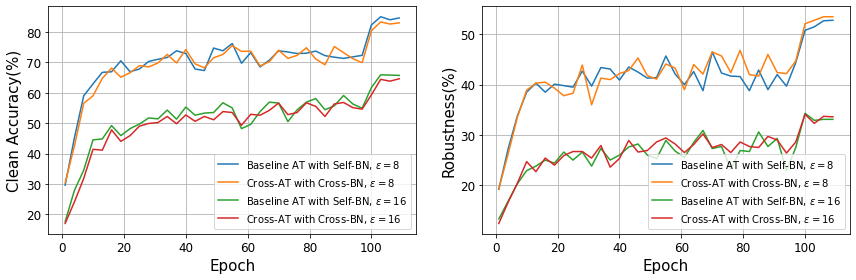}
  \caption{Clean accuracy and robustness (PGD10 Accuracy)  of Cross-AT during training. In Cross-AT,  the \textit{adversarial} samples are normalized by the BN statistics calculated by \textit{clean} samples. Interestingly, Cross-AT yield comparable robustness to original Self-BN(BN$_{adv}$). }
  \label{fig:cross_at}
\end{figure}

\begin{table}[!tbp]
\centering
\caption{Test accuracy of Cross-Hybrid-AT ($\epsilon=16/255$). In Cross-Hybrid-AT, the adversarial branch is normalized by BN$_{clean}$, and the clean branch is normalized by BN$_{adv}$. Experimental results show that Cross-Hybrid-AT achieves comparable results to Hybrid-AT with vanilla Dual BN.
}
\label{tab:cross_hybrid_at}
\resizebox{0.6\linewidth}{!}{
\begin{tabular}{ l |l | l | cccc}
\hline
Model &  Training  & Test  &Clean   & PGD-10 & AA \\
\hline 
Hybrid-AT & Dual BN &  BN$_{adv}$ &61.84 &  31.67 & 22.51  \\
Cross-Hybrid-AT &  Dual BN &   BN$_{clean}$ &59.56 &31.25 & 22.40\\
\hline 
Hybrid-AT & \multicolumn{2}{|c|}{Single BN} & 93.70 & 29.86 &  0.48 \\
\hline
\end{tabular}}
\end{table}

\section{On the BN induced misalignment} 
\label{sec:motivation}

In Hybrid-AT, the model is trained with two branches: a clean branch and an adversarial branch. These two branches share all model weights but are found to require independent BN modules, i.e., Dual BN~\citep{xie2019intriguing,xie2020adversarial}. At test time, only a single branch can be used by choosing either BN$_{adv}$ or BN$_{clean}$.
The adversarial branch (with BN$_{adv}$) is adopted in ~\citet{xie2019intriguing} for prioritizing high model robustness, while BN$_{clean}$ is adopted in ~\citet{xie2020adversarial} for only considering clean accuracy. 

\begin{wraptable}[8]{R}{0.4\textwidth}
\vspace{-23pt}
\centering
\caption{Test accuracy (\%)  of Hybrid-AT with Dual BN. BN$_{clean}$ leads to almost zero robustness under both perturbation budgets ($\epsilon$): 8/255 and 16/255. 
}
\label{tab:motivation_hybrid}
\resizebox{1.0\linewidth}{!}{  
\begin{tabular}{c| l| c c c }
\hline
$\epsilon$&Setups 
 & Clean  & PGD-10 & AA\\
\hline
8/255 &Dual BN (BN$_{adv}$) & 82.77 & 51.33 & 46.19 \\ 
&Dual BN (BN$_{clean}$)  & 94.91 &0.32  &0.10 \\ 
\hline 
16/255&Dual BN (BN$_{adv}$) & 61.84 & 31.67 & 23.14  \\ 
&Dual BN (BN$_{clean}$) &94.18 &0.00  &0.00 \\ 
\hline
\end{tabular}}
\end{wraptable}

However, swapping the BN during inference, i.e., adopting BN$_{clean}$ for robustness and BN$_{adv}$ for clean accuracy, leads to a significant performance drop. As shown in Table~\ref{tab:motivation_hybrid}, BN$_{clean}$ leads to almost zero robustness during inference. This interesting phenomenon inspires us to investigate the following question: \textit{will BN$_{clean}$ achieve robustness if it is trained with the adversarial branch, and vice versa?} To facilitate discussion of the above misalignment, we introduce a new term \textbf{Cross-BN} which refers to adopting BN$_{clean}$ for the adversarial branch or BN$_{adv}$ for the clean branch. With a similar terminology rule, BN$_{clean}$ for the clean branch or BN$_{adv}$ for the adversarial branch is termed as \textbf{Self-BN}.

\begin{wrapfigure}[12]{R}{0.45\textwidth}
\vspace{-20pt}
  \centering
 \includegraphics[width=1.0\linewidth]{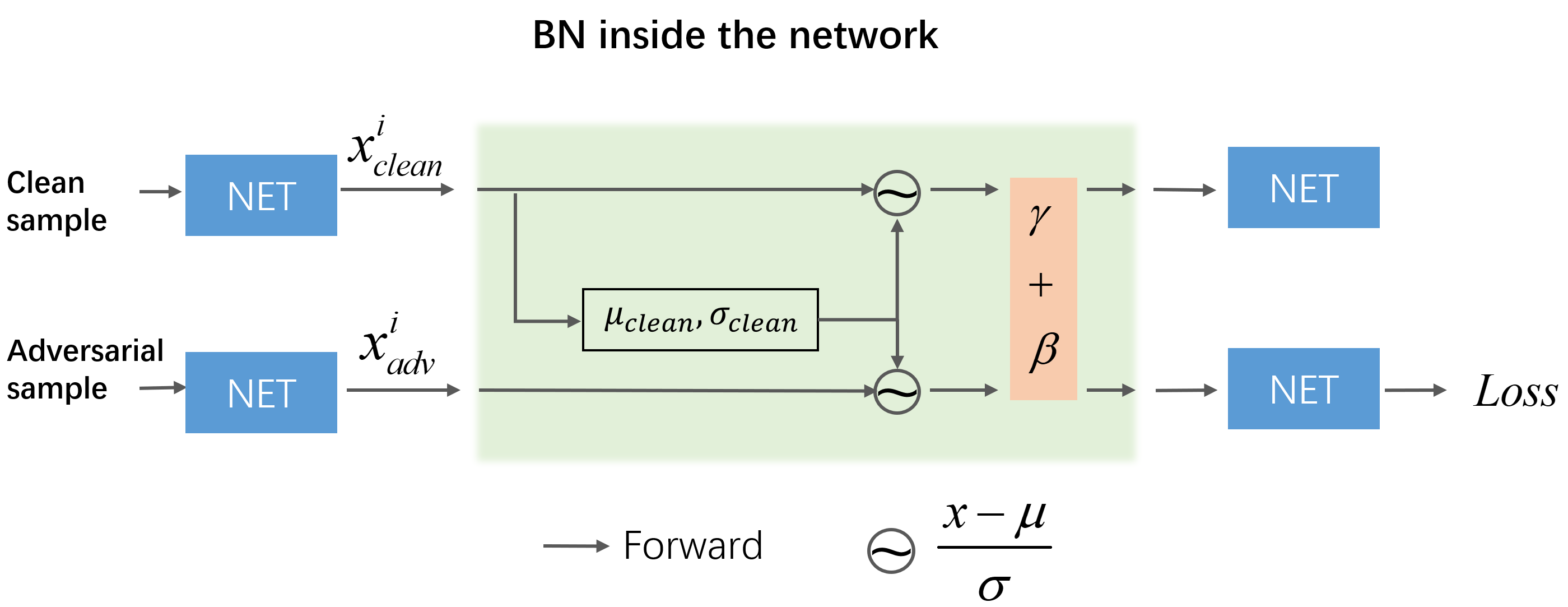}
  \caption{Cross-AT: Replacing BN$_{adv}$ with BN$_{clean}$ in the adversarial branch. The adversarial samples are normalized by the BN statistics calculated by clean samples.
  }
  \label{fig:oct_method}
\end{wrapfigure}
\textbf{Cross-AT: a preliminary investigation.} Before investigating Hybrid-AT with Cross-BN, we first investigate a setting where \textit{only} adversarial samples are used for model training. Note that it is adversarial branch, and the baseline model with a Self-BN adopts BN$_{adv}$. Cross-AT is conducted by replacing the default BN$_{adv}$ with a Cross-BN, i.e., BN$_{clean}$ (see Figure ~\ref{fig:oct_method}). Specifically, the adversarial samples are normalized by the BN statistics calculated by clean samples. It should be noted that in Cross-AT, the clean samples are used only for forward propagation to get the BN statistics, and the model weights are updated only by the adversarial branch. Interestingly, although the adversarial branch is normalized by BN${clean}$, Figure~\ref{fig:cross_at} shows that Cross-AT achieves comparable performance as the baseline model with Self-BN(BN$_{adv}$).

\textbf{Cross-Hybrid-AT: Hybrid-AT with Cross-BN.} Here, for the Dual BN in Hybrid-AT, we replace the default Self-BN with Cross-BN and term it Cross-Hybrid-AT. In Cross-Hybrid-AT, the adversarial branch is normalized by BN$_{clean}$, and the clean branch is normalized by BN$_{adv}$. As shown in Table~\ref{tab:cross_hybrid_at}, BN$_{clean}$ in Cross-Hybrid-AT achieves comparable results to BN$_{adv}$ in Hybrid-AT. The finding in Cross-Hybrid-AT is consistent with that in Cross-AT, which indicates that Cross-BN achieves comparable results to Self-BN.

\begin{table}[!tbp]
\centering
\caption{Test accuracy (\%) of untwining NS and AP in Dual BN. For NSs, 1 indicates mixed distribution and 2 indicates disentangled distribution for normalization. For  APs, 1 indicates a single set and 2 indicates double sets of APs.  The subscripts of AP$_{adv}$ and AP$_{clean}$ indicate the input data type used during training. Setup1 with two sets of APs achieves comparable results with Dual BN.
}
\label{tab:disentangling8}
\resizebox{0.78\textwidth}{!}{  
\begin{tabular}{ l|c c| c c c| c c c }
\hline
\multirow{2}{*}{Setups} & \multirow{2}{*}{NS}   & \multirow{2}{*}{AP} 
&\multicolumn{3}{c}{$\epsilon=8/255$}& \multicolumn{3}{|c}{$\epsilon=16/255$}\\ 
\cline{4-9}
&& & Clean  & PGD-10 & AA & Clean  & PGD-10 & AA\\
\hline 
Single BN & 1 & 1 & 88.06 & 49.75 & 7.03 & 93.70 & 29.86 &  0.48\\ 
Dual BN (BN$_{adv}$) & 2 & 2 & 82.77 & 51.33 & 46.19& 61.84 & 31.67 & 23.14  \\ 
Dual BN (BN$_{clean}$) & 2 & 2 & 94.91 &0.32  &0.10&94.18 &0.00  &0.00 \\ 
\hline
Setup1 (AP$_{adv}$) & 1  & 2 & 81.86 & 50.99 &44.63& 60.02 & 30.89 &23.43   \\ 
Setup1 (AP$_{clean}$) & 1  & 2 &94.74  &0.10  &0.04 &94.30  &0.00  &0.00 \\
Setup2 (NS$_{adv}$) & 2  & 1 & 85.49  & 49.39  &  42.96& 55.91 & 21.92 &10.64 \\
Setup2 (NS$_{clean}$) & 2  & 1 & 89.22  & 49.48  &  42.95& 86.35 & 1.08  & 0.00 \\
\hline
\end{tabular}}
\end{table}

\textbf{Implication of the above results.} As discussed above, training the model with Cross-BN leads to a comparable performance as with Self-BN in Hybrid-AT. However, this finding appears counter-intuitive considering the results of Hybrid-AT with Single BN. As shown in Table~\ref{tab:cross_hybrid_at}, Single BN leads to almost zero robustness (0.48\%) under AA attack. Note that a single BN is calculated by a mixture of clean and adversarial samples. If calculating BN statistics on either clean examples or adversarial examples can lead to high robustness, how come training on BN calculated on hybrid samples leads to an AA robustness close to zero? This finding also conflicts with the claims in ~\citet{xie2019intriguing,xie2020adversarial} that emphasize the importance of NS, and inspire us to investigate how Dual BN works in Hybrid-AT.

\section{Understanding how Dual BN works in training of Hybrid-AT}
\label{sec:how_dual_bn_works}

The failure of prior claims  ~\citet{xie2019intriguing,xie2020adversarial} to explain our observations in  Section ~\ref{sec:motivation} inspires us to investigate how Dual BN works in Hybrid-AT. On top of the single BN as a default case, Dual BN introduces an auxiliary BN component and causes two changes: (i) disentangling the mixture distribution for normalization statistics (NS) and (ii) introducing two sets of affine parameters (AP). To fully understand Dual BN in Hybrid-AT, we delve into its mechanisms.

\begin{wrapfigure}[15]{R}{0.5\textwidth}
\vspace{-10pt}
  \centering
 \includegraphics[width=0.9\linewidth]{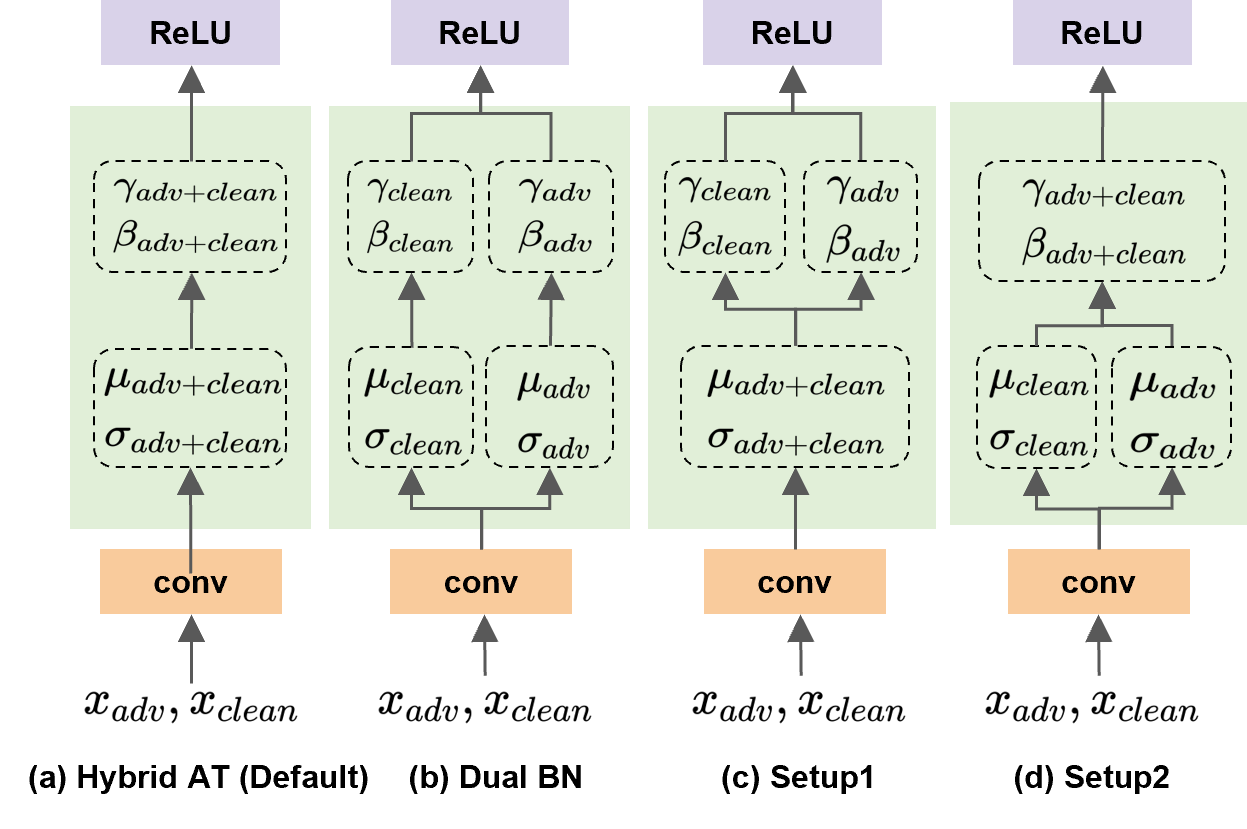}
\caption{Illustration of different BN setups for untwining NS and AP in Dual BN of Hybrid-AT.}
\label{fig:untwin_setup}
\end{wrapfigure}

\textbf{Untwining NS and AP in Dual BN.} As discussed above, compared with Hybrid-AT with Single-BN, Dual BN brings two effects: disentangled NSs and two sets of APs. To determine the influence of each effect on the model performance, we design two setups of experiments to include only one effect while excluding the other. In Setup1, we only include the effect of two sets of APs, by applying two different sets of APs ($\beta_{adv}$/$\gamma_{adv}$ and $\beta_{clean}$/$\gamma_{clean}$) in the adversarial and clean branches while using the default mixture distribution for normalization. In Setup2, we only include the effect of two sets of NSs by disentangling this mixture distribution with two different sets of NSs while making BN$_{clean}$ and BN$_{adv}$ share the same set of APs. The above setups of BNs are summarized in Figure~\ref{fig:untwin_setup} and we discuss the experimental results in Table~\ref{tab:disentangling8} as follows.

\textbf{Role of disentangled NS and AP.} As shown in Table~\ref{tab:disentangling8}, Dual BN (with BN$_{adv}$ during inference) brings significant robustness improvement over the Single BN baseline, which is consistent with findings in ~\citet{xie2019intriguing}. Interestingly, under the attack of PGD-10, their robustness gap is not significant, however, under AA, the Single BN achieves very low robustness (7.03\% and 0.48\% for $\epsilon=8/255$ and $16/255$, respectively). Moreover, Setup1 (AP$_{adv}$) achieves comparable robustness as that of Dual BN (BN$_{adv}$) for $\epsilon=8/255$ and $16/255$, suggesting two sets of APs alone achieve similar performance as Dual BN for yielding higher robustness (AP$_{adv}$) than single BN setting. 
The effect of two sets of NSs is more nuanced: for a small perturbation $\epsilon=8/255$, disentangling mixture distribution is beneficial for boosting the robustness under strong AA; for a large perturbation $\epsilon=16/255$, this benefit is less significant. This can be explained by the fact that training under $\epsilon=16/255$ is much harder than $\epsilon=8/255$.

\textbf{Conclusions.} Overall, we have two conclusions. First, two sets of AP  achieve comparable performance to Dual BN, aligning with the findings in ~\citet{rebuffi2023revisiting}. Moreover, our research extends beyond ~\citet{rebuffi2023revisiting}  by not only disentangling AP but also exploring the disentanglement of NS for a more thorough examination of Dual BN. Although not as effective as disentangling AP, disentangling NS can also achieve comparable robustness to Dual BN under certain conditions even against the strong AutoAttack, such as the small perturbation($\epsilon=8/255$). However, the benefit of disentangling NS narrows significantly for large perturbation.

\begin{figure}[!tbp]
  \centering
 \includegraphics[width=0.8\linewidth]{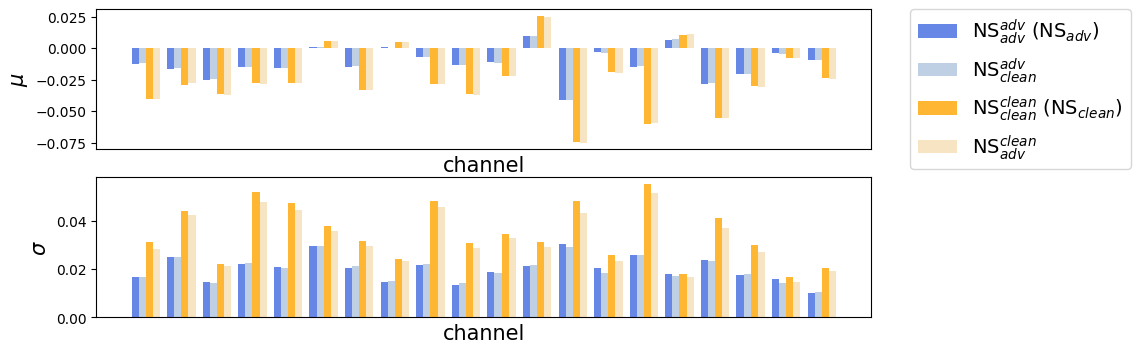}
  \caption{Visualization of normalization statistics (NS) by randomly choosing 20 channels and displaying the NS calculated with different APs. The superscript and subscript of NS refer to the AP  and input images when calculating NS, respectively. For example,  NS$_{clean}^{adv}$ is computed  on clean samples with AP$_{adv}$.  NSs calculated by the same AP are close to each other, such as NS$_{adv}^{adv}$ and NS$_{clean}^{adv}$ calculated by AP$_{adv}$, so is similar NS$_{clean}^{clean}$ and NS$_{adv}^{clean}$ calculated by AP$_{clean}$. 
  }
  \label{fig:mean_variance_visual}
\end{figure}

\section{On the domain gap between clean and adversarial samples}
\label{sec:two_domain} 

A model trained on a source domain performs poorly on a new target domain when there is a domain shift~\citep{daume2009frustratingly,sun2017correlation}. With BN as the target, it is common in the literature~\citep{li2016revisiting,benz2020revisiting,schneider2020improving,xie2019intriguing,xie2020adversarial} to indicate the domain gap by the difference of NS between two domains. In adversarial machine learning, prior work ~\citep{xie2019intriguing,xie2020adversarial,jiang2020robust}  perceive the adversarial domain as a new domain. Specifically, ~\citet{xie2019intriguing} highlights the adv-clean domain gap by visualizing the difference of NS in BN$_{adv}$ and BN$_{clean}$ (see Figure 5 of~\citep{xie2019intriguing}). The large domain gap visualized in ~\citep{xie2019intriguing} is somewhat in conflict with the finding in Section ~\ref{sec:how_dual_bn_works} that disentangling NS plays a less role for high performance. We further investigate the adv-clean domain gap for a comprehensive understanding.

\begin{figure}[!tbp]
  \centering
 \includegraphics[width=0.8\linewidth]{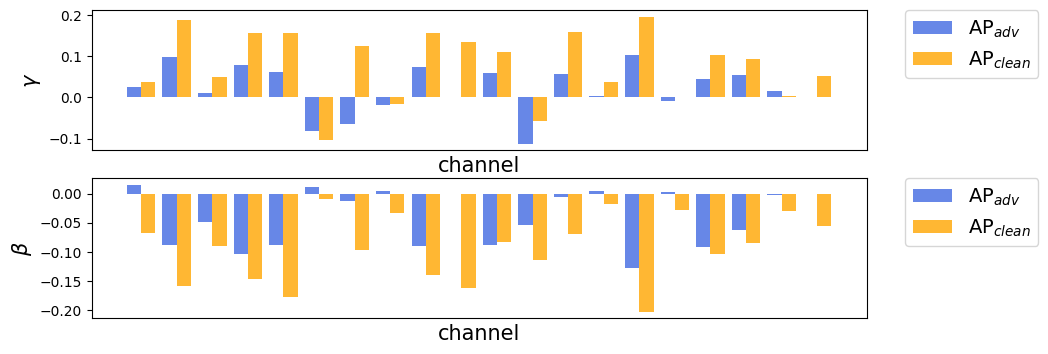}
  \caption{Visualization of affine parameters (AP). Randomly chose 20 channels for visualizing  AP$_{clean}$ and AP$_{adv}$.  There exists a gap between  AP$_{clean}$ and AP$_{adv}$.
  }
  \label{fig:beta_gamma_visual}
\end{figure}

\subsection{A hidden flaw leads to a misleading visualization}
\label{sec:flaw_visualize}

 With our analysis in Section.~\ref{sec:how_dual_bn_works}, we know that the AP in BN$_{clean}$ and BN$_{adv}$ are different. The clean branch and adversarial branch still have different weights, i.e., AP, even though the same set of convolutional filters are shared.  Therefore, the NS difference between BN$_{clean}$ and BN$_{adv}$ is characterized by two factors: (a) AP inconsistency and (b) different domain inputs. We discuss the influence of these two factors on the NS difference as follows.

 \textbf{Re-calibrated NS for disentangled analysis.} In the default setup of Dual BN, NS$_{clean}$ is calculated on clean samples with AP$_{clean}$, while NS$_{adv}$ is calculated on adversarial samples with AP$_{adv}$. In order to analyze the influence of different AP and domain inputs on the NS, we additionally calculate the NS on clean samples with AP$_{adv}$ (denoted as NS$_{clean}^{adv}$) and calculate the NS on adversarial samples with AP$_{clean}$ (denoted as NS$_{adv}^{clean}$). These two NS are termed \textbf{re-calibrated NS} since the AP and inputs are from different branches. Following NS$_{adv}^{clean}$ and NS$_{clean}^{adv}$ to indicate AP choice with the superscript and indicate sample choice with the subscript, we can also denote vanilla NS$_{clean}$ as NS$_{clean}^{clean}$ and denote NS$_{adv}$ as NS$_{adv}^{adv}$. Both NS$_{clean}^{clean}$ and NS$_{adv}^{adv}$ are  termed as \textbf{vanilla NS} for differentiation. Details of obtaining various NS is reported in Section~\ref{sec:appendix_setups} of the appendix.

 \textbf{A hidden flaw of NS visualization  in ~\citet{xie2019intriguing}.}   To exclude the influence of AP inconsistency, we intend to compare NS between clean and adversarial samples with the same AP (the superscript in NS). In other words, the domain gap is characterized by the difference between NS$_{clean}^{clean}$ and NS$_{adv}^{clean}$ or that between NS$_{adv}^{adv}$ and NS$_{clean}^{adv}$.  Following the procedures in ~\citet{xie2019intriguing}, we plot different types of NS in Figure~\ref{fig:mean_variance_visual} by randomly sampling 20 channels of the second BN layer in the first residual block. Fig.~\ref{fig:mean_variance_visual} shows that there exists a gap between NS$_{clean}^{clean}$ and NS$_{adv}^{adv}$, which is consistent with the findings in ~\citet{xie2019intriguing}. Moreover, there are two other observations from Figure~\ref{fig:mean_variance_visual}. First, if we fix the input samples and calculate NS with different AP, there exists a large gap, i.e., the gap between NS$_{clean}^{adv}$ and NS$_{clean}^{clean}$, as well as the gap between NS$_{adv}^{adv}$ and NS$_{adv}^{clean}$.  Second,  those NSs with the same APs are very close to each other: NS$_{adv}^{adv}$ and NS$_{clean}^{adv}$ are very similar to each other, and the same applies to NS$_{adv}^{clean}$ and NS$_{clean}^{clean}$.   
 We report the visualization results of AP in Figure~\ref{fig:beta_gamma_visual} for comparison, which shows a significant gap between AP$_{clean}$ and AP$_{adv}$.

\textbf{Conclusions.} We point out a flaw in ~\citet{xie2019intriguing} that the large adv-clean gap visualized in ~\citet{xie2019intriguing} is caused by the AP consistency. When adopting the same AP, the adv-clean domain gap significantly narrows. Our investigations suggest that the visualization and conclusions in ~\citet{xie2019intriguing} might convey a misleading message. Our findings update the understanding on the adv-clean domain gap.

\subsection{Adv-clean domain gap is not as large as expected}
\label{sec:small_domain_gap}

\begin{SCfigure}
    \centering
    \includegraphics[width=0.35\textwidth]{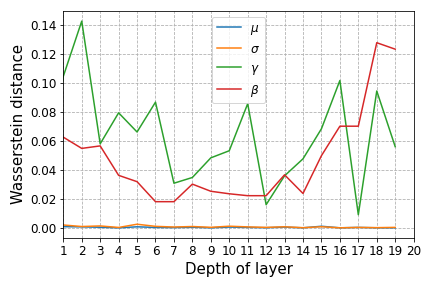}
    \caption{Layer-wise discrepancy visualization. For all layers, there exists a large distance (higher than zero) between AP$_{clean}$ and AP$_{adv}$, see  $\gamma$ and $\beta$ in the figure. However, with the same AP$_{adv}$, the gap between NS$_{adv}^{adv}$ and NS$_{clean}^{adv}$  stays almost zero in all layers, see  $\mu$ and $\sigma$ in the figure.
    }
    \label{fig:distance_layers_training}
\end{SCfigure}

\textbf{Quantitative measurement of domain gap.} Figure~\ref{fig:mean_variance_visual} investigates the adv-clean domain gap qualitatively. For a quantitative comparison, we measure the Wasserstein distance between clean and adversarial branches in different layers in Figure~\ref{fig:distance_layers_training}. As shown in Figure~\ref{fig:distance_layers_training}, the Wasserstein distance of NS between clean and adversarial branches is much smaller than the difference of AP for a certain layer. This finding is consistent with that in Figure~\ref{fig:mean_variance_visual} and Figure~\ref{fig:beta_gamma_visual}.

\begin{wraptable}[5]{ht}{0.45\textwidth}
\vspace{-25pt}
\centering
\caption{Test accuracy (\%) under random noise and adversarial perturbation during inference.}
\label{tab:noise_adversarial}
\resizebox{1.0\linewidth}{!}{
\begin{tabular}{l|cccc}
\hline
Noise/perturbation Size & 0 & 8/255 & 16/255 \\
\hline
Random noise & 94.0 & 92.7 & 86.6 \\
Adversarial perturbation & 94.0 & 0.00 & 0.00 \\
\hline
\end{tabular}}
\end{wraptable}

\textbf{Adv-clean versus noisy-clean domain gap.}  As suggested in ~\citet{benz2020revisiting,schneider2020improving}, noisy samples (images corrupted by random noise) can be seen as a domain different from clean samples. Adversarial perturbation is a \textit{worst-case} noise for attacking the model. Taking a ResNet18 model trained on clean samples for example, we report the performance under adversarial perturbation and random noise (with the same magnitude) in Table~\ref{tab:noise_adversarial}. 
As expected,  the model accuracy drops to zero with adversarial perturbation. Under random noise of the same magnitude, we find that the model performance only drops by a small margin. Given that the influence of adversarial perturbation on the model performance is significantly larger than that of random noise, it might be tempting to believe that the adversarial-clean domain gap is much larger than noisy-clean domain gap.

\begin{SCfigure}
    \centering
\includegraphics[width=0.35\linewidth]{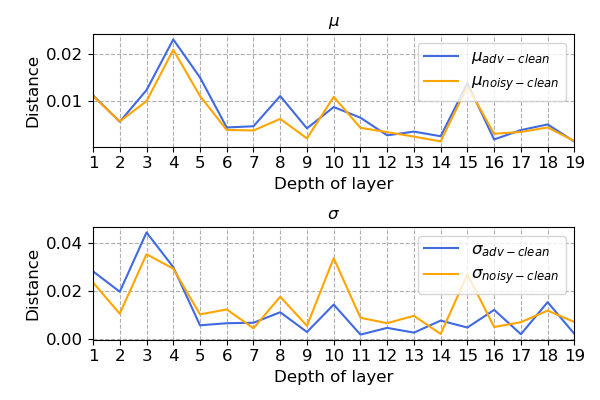}
\caption{Visualization of adversarial-clean domain gap and noisy-clean domain gap (perturbation/noise magnitude is set to $16/255$).}
\label{fig:adv_noisy_16_single_img}
\end{SCfigure}

With Wasserstein distance of NS between different domains as the metric, we compare the adversarial-clean domain gap with noisy-clean counterpart on the above ResNet18 model trained on clean samples, as shown in Figure~\ref{fig:adv_noisy_16_single_img}. The perturbation and noise magnitude are set to $16/255$.
Interestingly, we observe that there is no significant difference between the adversarial-clean domain gap and noisy-clean counterpart.

% In other words, 

\textbf{Conclusions.} Based on the visualization and quantitative results, we reveal that the adversarial-clean domain gap is not as large as many might expect, considering the strong performance drop caused by adversarial perturbation.

\subsection{Interpreting Hybrid-AT from a two-task perspective}
\label{sec:two_task}

\begin{wraptable}[8]{ht}{0.50\textwidth}
\vspace{-22pt}
\centering
\caption{Test accuracy of Hybrid-AT with Dual linear, $\epsilon=16/255$. Hybrid-AT with Dual Linear achieves a similar trend and comparable results with Dual BN. 
}
    \label{tab:hybrid_two_linear}
    \resizebox{1.0\linewidth}{!}{
    \begin{tabular}{ l|l|ccc}
    \hline
    Setups &Branch    & Clean & PGD10 & AA\\ 
    \hline
     \multirow{2}{*}{Dual BN}  & BN$_{adv}$   & 61.84  &31.67&22.51\\
      & BN$_{clean}$ &  94.18 &0.00  &0.00 \\ 
    \hline
     \multirow{2}{*}{Dual Linear} & Linear$_{adv}$  &  60.72 &28.84 & 16.50\\ 
    & Linear$_{clean}$ & 91.43 & 2.21  &1.30 \\  
    \hline
    \end{tabular}}
\end{wraptable}

\textbf{New empirical foundation for Hybrid-AT improvement.} Prior work ~\citep{xie2019intriguing} takes the two-domain hypothesis as the empirical foundation for applying Dual BN in Hybrid-AT. Our investigations in Section~\ref{sec:flaw_visualize} and Section~\ref{sec:small_domain_gap} identify a flaw in the two-domain hypothesis, and point out that the adv-clean domain gap is not as large as expected. These findings suggest discarding the two-domain hypothesis in Hybrid-AT improvement. Instead, we propose a new empirical foundation for Hybrid-AT improvement: the two-task hypothesis. Intuitively, with the two branches in Hybrid-AT, the model weights are trained for two tasks: one for clean accuracy and the other for robustness.
A common approach for handling two tasks with a shared backbone is to make the top layers unshared. Here, we experiment with a shared encoder of single BN but with dual linear classifiers. The results in Table~\ref{tab:hybrid_two_linear} show that Dual Linear results in similar behavior as Dual BN, validating the two-task hypothesis. ~\citet{rebuffi2023revisiting} proposes to use adapters for each type of input in Hybrid-AT model, which matches the classification performance of Dual BN with significantly fewer parameters. The success of adapters in ~\citet{rebuffi2023revisiting} also validates our two-task hypothesis.

\begin{wraptable}[8]{ht}{0.50\textwidth}
\vspace{-20pt}
\centering
\caption{Test accuracy of Hybrid-AT (Single BN) with KL loss, $\epsilon=16/255$. 
}
\label{tab:hybrid_single_with_kl}
\resizebox{1.0\linewidth}{!}{
\begin{tabular}{ l|ccc}
\hline
Setups      & Clean & PGD10 & AA\\ 
\hline
Single BN  & 93.70 & 29.86 &  0.48\\ 
Single BN (with KL loss) & 68.86 & 33.61 & 23.60\\ 
\hline
Dual BN (BN$_{adv}$)   & 61.84 &31.67 &22.51\\
Dual BN (BN$_{clean}$) & 94.18 &0.00  &0.00 \\ 
\hline
\end{tabular}}
% \end{table}
\end{wraptable}
\textbf{A unified framework for Hybrid-AT improvements.}
Our new perspective on Hybrid-AT enables alternative solutions to mitigate the two-task conflict without resorting to two sets of APs. Here, we experiment with including an additional regularization loss, which is introduced to minimize the gap between two tasks. As a concrete example, we add a KL loss on the basic loss of Hybrid-AT in Eq~\ref{eq:hybrid}. The extra loss is designed to explicitly minimize the discrepancy between the outputs of adversarial and clean branches. Interestingly, this simple change improves the AA result of Single BN significantly from 0.48\% to 23.60\%. Compared to BN$_{adv}$ (the default branch during inference), Hybrid-AT with KL loss achieves superior performance on both clean accuracy and robustness. 
Interestingly, the Hybrid-AT with KL loss reminds us of another AT framework termed Trades-AT~\citep{zhang2019theoretically}, which is also trained on hybrid samples and has a KL loss. This might provide an explanation for the effectiveness of Trades-AT~\citep{zhang2019theoretically} by analyzing the KL term. Admittedly, KL loss on the output is just a naive attempt, but its promising result invites future works to explore other solutions.

\textbf{Conclusions.} Our investigations on the adv-clean domain gap suggest discarding the prior two-domain hypothesis. Instead,  we propose a two-task hypothesis as the new empirical foundation for Hybrid-AT improvement. The two-task hypothesis also serves as a unified framework that collaborates different methods together, such as Dual BN, adapters in ~\citet{rebuffi2023revisiting} and Trades-AT~\citep{zhang2019theoretically} that seem to be unrelated at first sight.

\subsection{AP characterizes robustness in test-time}
 
Investigation on BN in test-time has been a widely discussed topic in other areas~\citep{li2016revisiting,benz2020revisiting,schneider2020improving}. Different from Section ~\ref{sec:how_dual_bn_works} and ~\citet{rebuffi2023revisiting} that investigates model training, we further investigate how Dual BN works in test-time. Specifically, we compare the robustness with various NS and AP combinations, as reported in Table~\ref{tab:hybrid_finegrained_eval}. Details of obtaining various NS are reported in Section~\ref{sec:appendix_setups} of the appendix.

\textbf{Robustness with various NS and AP pairs.}
Table~\ref{tab:hybrid_finegrained_eval} shows that given AP$_{adv}$,  re-calibrated NS$_{clean}^{adv}$ achieves a robustness of 51.75\%, which is comparable to 51.33\% with NS$_{adv}^{adv}$. Note that the only difference between NS$_{clean}^{adv}$ and NS$_{adv}^{adv}$ is that they are calculated by clean and adversarial samples, respectively. Moreover, given AP$_{clean}$, both  NS$_{adv}^{clean}$ and NS$_{clean}^{clean}$ yield  zero robustness. The results of swapping NS$_{clean}^{clean}$ and NS$_{adv}^{adv}$ when AP is fixed is also given in Table~\ref{tab:hybrid_finegrained_eval} for comparison. We find that directly replacing NS$_{adv}^{adv}$ with NS$_{clean}^{clean}$ in BN$_{adv}$ (Setup1) results in lower robustness (17.1\%) than original BN$_{adv}$ (51.33\%).

\textbf{Conclusions.} We conclude that AP characterizes the large robustness gap between BN$_{clean}$ and BN$_{adv}$ during inference. When AP is consistent for both NS computation and robustness evaluation, the robustness gap between the NS calculated on clean or adversarial samples is limited.

\begin{table}[!tbp]
\centering
\caption{Evaluation results of various NS and AP pairs, $\epsilon=16/255$. During inference, 
 \textbf{re-calibrated} NS  achieves comparable performance to the default setting.} 
\label{tab:hybrid_finegrained_eval}
\resizebox{0.65\linewidth}{!}{
\begin{tabular}{ l|l|l  l|   cc |cc}
\hline
 \multirow{2}{*}{} & Setups&\multirow{2}{*}{NS}    & \multirow{2}{*}{AP}  & \multicolumn{2}{c|}{$\epsilon=8/255$} & \multicolumn{2}{c}{$\epsilon=16/255$}\\ 
 \cline{5-8}
 &&&&PGD10 & AA& PGD10 & AA \\
 \hline
 \multirow{2}{*}{Default} & BN$_{adv}$  & NS$_{adv}^{adv}$   &  AP$_{adv}$  & 51.33& 46.19 &31.67&22.51\\
  &  BN$_{clean}$ & NS$_{clean}^{clean}$   & AP$_{clean}$ &   0.32 & 0.10&0.00&0.00\\ 
\hline 
\multirow{2}{*}{Swap} & Setup1 &  NS$_{clean}^{clean}$  & AP$_{adv}$ &17.1&9.16&10.02&9.80\\ 
 & Setup2 &  NS$_{adv}^{adv}$  & AP$_{clean}$ & 0.00&0.00&0.45&0.00 \\
\hline
 \multirow{2}{*}{Re-calibration} & Setup3 & NS$_{clean}^{adv}$  & AP$_{adv}$ &51.75&46.55& 32.73&24.40\\ 
&Setup4&  NS$_{adv}^{clean}$   & AP$_{clean}$ &0.00&0.00& 0.00&0.00 \\  
\hline
\end{tabular}}
\end{table}

\section{Related work}

\textbf{Adversarial training.} Since the advent of Classical-AT~\citep{madry2017towards} and Hybrid-AT~\citep{xie2019intriguing,xie2020adversarial}, numerous works have attempted to improve AT from various perspectives. From the data perspective, ~\citet{uesato2019labels,carmon2019unlabeled,zhang2019bottleneck} have independently shown that unlabeled data can be used to improve the robustness. From the model perspective, AT often benefits from the increased model capacity of models~\citep{uesato2019labels,xie2019intriguing}. ~\citet{xie2020smooth,pang2020bag,gowal2020uncovering} have investigated the influence and suggested that a smooth activation function, like parametric softplus, is often but not always~\citep{gowal2020uncovering} helpful for AT. Another branch of studies aims to improve the training efficiency of adversarial training based on PGD attack, termed as FAST AT~\citep{de2022make,jia2022boosting,park2021reliably,wong2020fast,andriushchenko2020understanding,jia2022prior}. Specifically, FGSM attack is adopted in ~\citet{wong2020fast,andriushchenko2020understanding,de2022make} to replace PGD attack during training, which achieves promising robustness with catastrophic overfitting problem tackled.

\textbf{Dual BN in AT.}  Prior work~\citep{xie2020adversarial} shows that adversarial samples can be used to improve recognition (accuracy) by adversarial training where adversarial samples are normalized by an independent BN$_{adv}$. 
Moreover,~\citet{xie2019intriguing} has shown that adding clean images in adversarial training (AT) can significantly decrease robustness performance, where such negative effects can be alleviated to a large extent by simply normalizing clean samples with an independent BN$_{clean}$. Inspired by their finding, ~\citet{jiang2020robust} also adopts Dual BN in adversarial contrastive learning, showing that single BN performs significantly worse than Dual BN. Beyond Dual BN, triple BN has been attempted in ~\citet{fan2021does} for incorporating another adversarial branch. ~\citet{wang2021augmax} has also combined Dual BN with Instance Normalization to form Dual batch-and-Instance Normalization for improving robustness.  Prior work ~\citep{xie2019intriguing} interprets the necessity of Dual BN from the perspective of an inherent large adversarial-clean domain gap. By contrast, ~\citet{rebuffi2023revisiting} demonstrates that separate batch statistics are not necessary for Hybrid-AT and it is sufficient to use adapters with few domain-specific parameters for each type of input. Extensive experimental results in ~\citet{rebuffi2023revisiting} show that the proposed model matches Dual BN's performance and the adversarial model soups perform better on ImageNet variants than the advanced masked auto-encoders.  Our investigations extends the scope of ~\citet{rebuffi2023revisiting} by reporting new findings on previously unaddressed topics, including but not limited to the role of disengaged NS in model training and the new understanding of adversarial-clean domain gap. Our two-task hypothesis extends the adapter method proposed in ~\citet{rebuffi2023revisiting} by  not only underpinning the adapter method with empirical evidence but also serving as a foundation for various methods utilizing domain-specific trainable parameters. This hypothesis also links the adapter method to broader Hybrid-AT enhancement strategies, such as Trades-AT~\citep{zhang2019theoretically}. Additionally, the adapter's success in ~\citet{rebuffi2023revisiting} reinforces our hypothesis, highlighting how our work complements and expands upon the contributions of ~\citet{rebuffi2023revisiting}.

\textbf{BN applications beyond AT.} Prior work investigates BN 
 in various fields beyond AT, such as for adversarial transferability~\citep{benz2021batch,dong2022random}. ~\citet{benz2021batch} investigates  BN from the non-robust feature perspective. Specifically, ~\citet{benz2021batch} empirically reveals that  BN shifts a model towards being more dependent on the non-robust features. Based on this finding,  ~\citet{benz2021batch} suggests strategies like removing BN or early stopping during the training of substitute models to improve adversarial transferability.  Another work ~\citet{dong2022random} provides both empirical and theoretical evidence which shows that the upper bound of adversarial
transferability is influenced by the types and parameters of normalization layers. Based on this observation, ~\citep{dong2022random} proposes a Random Normalization Aggregation (RNA) module to replace original normalization layers and create a combination of different sampled normalization. Extensive experiments demonstrate that the proposed RNA module achieves superior performance on different datasets and models.  Another branch of work adopts Dual BN in domain adaptation. AdaBN ~\citep{li2016revisiting} leverages different statistics for two domains but loses source domain information by using only target domain statistics during inference. DSBN~\citep{chang2019domain} introduces a separate BN branch for unsupervised domain adaptation, extendable to multisource scenarios. \citet{huang2023reciprocal} proposes reciprocal normalization that structurally aligns the source and target domains by conducting reciprocity across domains. In this work, we mainly focus on investigating Dual BN in the Hybrid-AT.

\section{Conclusion}

We experiment with Cross-AT and demonstrate the compatibility of clean samples' BN statistics with the adversarial branch, which inspires us to doubt the claims of prior work for justifying the necessity of Dual BN in Hybrid AT.
We investigate the effect of disentangled NS and AP on training a Hybrid-AT model, leading to a thorough refutation of prior claims about the significance of NS.  Our work further identifies a visualization flaw of the prior two-domain hypothesis, and points out that the adversarial-clean domain gap is not as large as expected.  In addition, we propose a new interpretation of Hybrid-AT with Dual BN from the two-task perspective. Finally, we investigate different types of NS and AP in test-time, revealing that AP characterizes robustness during inference. Our study provides a comprehensive understanding of Dual BN as well as the adversarial examples. 

\subsubsection*{Acknowledgments}

This work was conducted by Center for Applied Research in Artificial Intelligence (CARAI) grant funded by DAPA and ADD (UD230017TD).

\bibliography{bib_mixed,bib_local}
\bibliographystyle{tmlr}

\appendix

\section{Appendix}

\subsection{Experimental Setups}
\label{sec:appendix_setups}

\textbf{Experimental setups.} We train and evaluate all models in the paper using the same setups following ~\citet{pang2020bag}. In this work, we perform experiments on CIFAR10~\citep{krizhevsky2009learning,andriushchenko2020understanding,zhang2022decoupled} with ResNet18~\citep{andriushchenko2020understanding,targ2016resnet,wu2019wider,li2016demystifying,zhang2022decoupled} and follow the suggested training setups in ~\citet{pang2020bag} unless specified. Specifically, we train the model for 110 epochs. The learning rate is set to 0.1 and decays by a factor of 0.1 at the epoch 100 and 105. We adopt an SGD optimizer with weight decay $5\times10^{-4}$. For generating adversarial examples during training, we use $\ell_{\infty}$ PGD attack with 10 iterations and step size $\alpha=2/255$. For the perturbation constraint, $\epsilon$ is set to $\ell_{\infty}$ $8/255$~\citep{pang2020bag} or $16/255$~\citep{xie2019intriguing}. Following~\citet{pang2020bag}, we evaluate the model robustness under PGD-10 attack (PGD attack with 10 steps) and AutoAttack (AA)~\citep{croce2020reliable}.

\textbf{Re-calibrate NS.} The NS$_{clean}^{clean}$ and NS$_{adv}^{adv}$ correspond to original Dual BN's NS$_{clean}$ and NS$_{adv}$, respectively. For re-calibrated NS, we simulate vanilla NS calculations, specifying the input domain following the subscript and AP following the superscript. For example, NS$_{clean}^{adv}$ is computed by processing clean images through the branch with AP$_{adv}$. To simulate the computation of running means in the original Dual BN, we forward the samples for multiple epochs for  converged results.  In Table ~\ref{tab:hybrid_finegrained_eval}, we first calculate various types of NS and then evaluate robustness by combining NS with AP.

\textbf{Training details of models in Section 5.3.} For the Dual Linear model, we add two linear layers above its penultimate layer and train it using the Hybrid loss outlined in Equation 1. For models with regularization loss, we train using a single BN and augment  the original Hybrid-AT loss with an additional KL-divergence between the predictions on adversarial and clean inputs.

\subsection{Further investigations beyond BN}

Inspired by finding that two sets of AP can achieve comparable results to Dual BN, we further investigate whether this holds in cases beyond BN where disentangling NS is not applicable. For example, layer normalization (LN) adopts sample-wise NS, and therefore it is not applicable to disentangle distribution-wise NS between two domains. We experiment with dual AP on ResNet  with LN and the results are reported in Table~\ref{tab:multiple_norm}. We observe that with Dual AP, LN  performs similarly with BN in either setup (b) and (c) in Figure~\ref{fig:untwin_setup}. We also investigate other normalization methods and model architectures (such as ViT) in Table~\ref{tab:multiple_norm}. Table 8 shows that across various normalization methods and architectures, two AP sets achieve comparable performance to Dual BN while single AP set fails to achieves high robustness against AA.

\begin{table}[!htbp]
\centering
\caption{Effect of dual AP on various types of normalizations and models($\epsilon=16/255$), where LN, GN and IN denote Layer Normalization, Group Normalization and Instance Normalization, respectively. }
\label{tab:multiple_norm}
\resizebox{0.5\linewidth}{!}{
\begin{tabular}{ l|l|l|l| ccc}
\hline
 Norm  & Norm &Setups  & Branch   & Clean & PGD10 & AA\\ 
\hline
ResNet &BN &  Single BN  & /& 93.70 & 29.86 &  0.48\\ 
\cline{3-7} 
& & Dual BN  & BN$_{adv}$   & 61.84 &31.67 &22.51\\
& & Dual BN & BN$_{clean}$ & 94.18 &0.00  &0.00 \\ 
\cline{2-7} 
 & LN & Single AP & /& 75.12 & 18.81 & 11.80\\ 
\cline{3-7} 
&     &  Dual AP & AP$_{adv}$ & 62.56&26.98  & 16.90  \\
&     &  Dual AP & AP$_{clean}$ &88.41& 0.00&0.00 \\ 
\cline{2-7} 
 & GN& Single AP & /& 81.85 & 21.94 & 14.50\\ 
\cline{3-7} 
 & &  Dual AP & AP$_{adv}$ &70.27 & 29.36 &18.30  \\
 & &  Dual AP & AP$_{clean}$ & 91.82 & 0.00 & 0.00\\ 
\cline{2-7} 
& IN& Single AP & /&92.55 & 23.06 & 1.20 \\ 
\cline{3-7} 
& &  Dual AP & AP$_{adv}$ &52.29 & 25.27 & 16.10  \\
& &  Dual AP & AP$_{clean}$  &92.35 & 0.00 &  0.00\\ 
\hline
ViT  & LN  & Single AP   &/    & 92.21 & 33.60   & 1.84   \\
\cline{3-7} 
     &    & Dual AP  & $AP_{clean}$   & 58.02 & 30.08   & 12.44  \\
     &    & Dual AP  &$AP_{adv}$  & 91.60 & 0.00    & 0.00   \\
\hline
\end{tabular}}
\end{table}

\end{document}